\documentclass[letterpaper]{article} % DO NOT CHANGE THIS
\usepackage{aaai25}  % DO NOT CHANGE THIS
\usepackage{times}  % DO NOT CHANGE THIS
\usepackage{helvet}  % DO NOT CHANGE THIS
\usepackage{courier}  % DO NOT CHANGE THIS
\usepackage[hyphens]{url}  % DO NOT CHANGE THIS
\usepackage{graphicx} % DO NOT CHANGE THIS
\usepackage{natbib}  % DO NOT CHANGE THIS AND DO NOT ADD ANY OPTIONS TO IT
\usepackage{caption} % DO NOT CHANGE THIS AND DO NOT ADD ANY OPTIONS TO IT
\usepackage{algorithm}
\usepackage{algorithmic}

\usepackage{array, booktabs}
\usepackage{amsmath, amsfonts}
\usepackage{newfloat}
\usepackage{listings}
\DeclareCaptionStyle{ruled}{labelfont=normalfont,labelsep=colon,strut=off} % DO NOT CHANGE THIS
\floatstyle{ruled}
\newfloat{listing}{tb}{lst}{}
\floatname{listing}{Listing}
\title{Towards Efficient Object Re-Identification with A Novel Cloud-Edge Collaborative Framework}
\author{
    Chuanming Wang\equalcontrib,
    Yuxin Yang\equalcontrib,
    Mengshi Qi,
    Huanhuan Zhang,
    Huadong Ma\thanks{\noindent Corresponding author: Huadong Ma.}
}

\affiliations{
    The State Key Laboratory of Networking and Switching Technology\\ Beijing University of Posts and Telecommunications
}

\begin{document}

\maketitle

\begin{abstract}
    Object re-identification (ReID) is committed to searching for objects of the same identity across cameras, and its real-world deployment is gradually increasing. Current ReID methods assume that the deployed system follows the centralized processing paradigm, i.e., all computations are conducted in the cloud server and edge devices are only used to capture images. As the number of videos experiences a rapid escalation, this paradigm has become impractical due to the finite computational resources in the cloud server. Therefore, the ReID system should be converted to fit in the cloud-edge collaborative processing paradigm, which is crucial to boost its scalability and practicality. However, current works lack relevant research on this important specific issue, making it difficult to adapt them into a cloud-edge framework effectively. In this paper, we propose a cloud-edge collaborative inference framework for ReID systems, aiming to expedite the return of the desired image captured by the camera to the cloud server by learning the spatial-temporal correlations among objects. In the system, a Distribution-aware Correlation Modeling network (DaCM) is particularly proposed to embed the spatial-temporal correlations of the camera network implicitly into a graph structure, and it can be applied 1) in the cloud to regulate the size of the upload window and 2) on the edge device to adjust the sequence of images, respectively. Notably, the proposed DaCM can be seamlessly combined with traditional ReID methods, enabling their application within our proposed edge-cloud collaborative framework. Extensive experiments demonstrate that our method obviously reduces transmission overhead and significantly improves performance. 
\end{abstract}

\begin{links}
    \link{Code}{https://github.com/bupt-wcm/AAAI-DaCM.git}
\end{links}

\section{Introduction}
\label{sec:intro}

Object re-identification (ReID)~\cite{DBLP:conf/iccv/He0WW0021, DBLP:conf/mm/HeLLLCM23, DBLP:conf/aaai/LiSL23, DBLP:conf/cvpr/0004GLL019,DBLP:conf/cvpr/Huynh21, DBLP:journals/tnn/GeZCZWL24, DBLP:conf/cvpr/ChenWLDB21, DBLP:journals/tomccap/FanZYY18, DBLP:journals/tmm/FuCWQM24, DBLP:journals/tip/QiQYWL21, DBLP:conf/mm/QiWL17, DBLP:journals/tmm/LiuLMM18} aims to retrieve specific objects captured by non-overlapping cameras, which usually serves as a fundamental task in the field of multimedia processing. 
It can facilitate users in searching objects accurately across diverse scenes and views, significantly alleviating manual overhead in visual surveillance. 
With the increasing demands, ReID systems have been widely deployed in various real-world scenarios for vehicle or person searching, playing an important role in traffic monitoring, safety management, etc. 
Therefore, an expanding number of innovative technologies have been introduced to promote the accuracy of ReID system continuously, including establishing elaborate feature extractors~\cite{DBLP:conf/cvpr/0004GLL019, DBLP:conf/cvpr/Huynh21,DBLP:conf/iccv/He0WW0021, DBLP:conf/aaai/LiSL23}, developing data transmission schemes~\cite{DBLP:conf/ieeesec/JainZZAJSBG20}, and designing inference strategies~\cite{zhong2017rerank}.

\begin{figure}[t]
    \centering
        \includegraphics[width=0.95\linewidth]{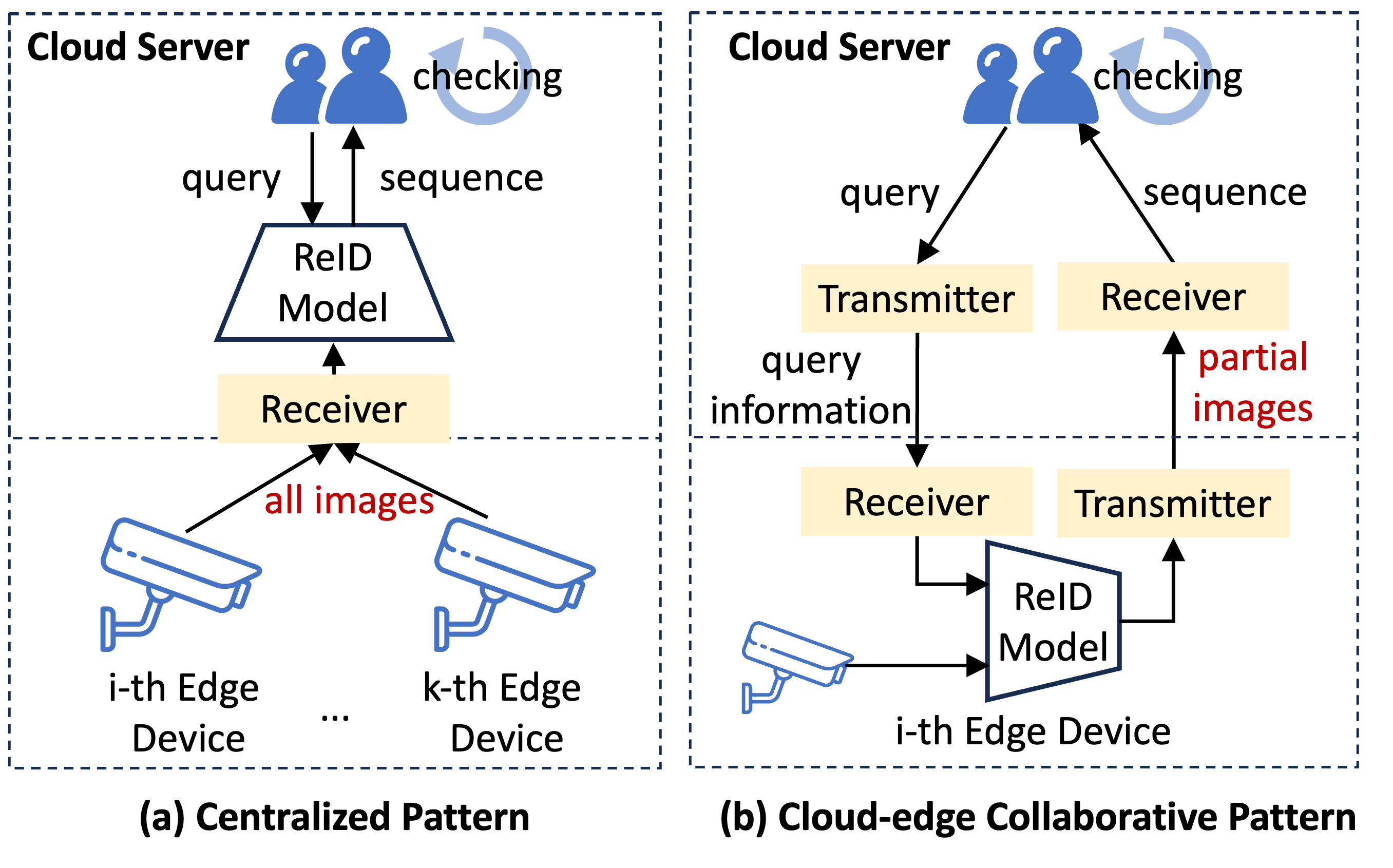}
        \caption{Illustration of the difference between centralized and cloud-edge collaborative patterns for ReID systems. 
        }~\label{fig:comp}
\end{figure}

Due to its intrinsic cross-scene nature, a ReID system typically consists of a central cloud server, multiple edge devices (such as cameras), and a communication network for transmitting images and associated data. Previous ReID methods typically follow a centralized processing pattern, where, as illustrated on the left of Fig.~\ref{fig:comp}, edge devices merely capture images and upload \textbf{all images} to the cloud server via the connected network. The cloud server then utilizes a deep neural network to extract features and compute similarities between the query and returned images. However, with the rapid proliferation of cameras, this processing pattern imposes excessive strain on the communication network’s bandwidth and the cloud server's computing and storage capacities, leading to significant service delays and a compromised user experience. Consequently, in line with current technological trends and driven by the advancement of device computing power~\cite{DBLP:journals/sensors/AngelRVSH22}, as depicted on the right of Fig.~\ref{fig:comp}, the ReID system should seamlessly integrate into a cloud-based collaborative framework. Feature extraction should occur locally at the edge device, with \textbf{partial images} being uploaded to the cloud server based on the ReID model’s outputs, thereby alleviating the burden on network communication and cloud computing.

Some previous methods~\cite{DBLP:journals/corr/abs-2008-11560,DBLP:conf/mm/Zhuang0Z21,DBLP:conf/cvpr/JiangXLZ23} want to establish new-style cloud-edge collaborative frameworks, but they primarily concentrate on the training phase. For instance, FedReID~\cite{DBLP:journals/corr/abs-2008-11560} and FedUReID~\cite{DBLP:conf/mm/Zhuang0Z21} embed the federated learning into ReID system, delving into strategies to exploit distributed data to continuously optimize the feature extractor, thereby enhancing search accuracy. Besides, some works~\cite{DBLP:conf/cvpr/JiangXLZ23} propose to deploy a segment of the deep neural network to the edge devices, mitigating the computing cost of the cloud server, although it still necessitates a substantial amount of data transmission. We argue that the inference phase also holds greater significance for a practical ReID system, and a meticulous scheme should be developed to fully leverage the advantages of both the cloud server and edge devices. Therefore, in contrast to previous methods, we introduce a pioneering cloud-edge collaborative ReID system that places a heightened emphasis on optimizing the efficiency and effectiveness of the inference process, a domain that has been under-explored in existing research.

For a basic cloud-edge collaborative inference pipeline of the ReID system, when the user requests to search one certain object, one query image and its auxiliary information (denoted as \textit{query} in Fig.~\ref{fig:comp}) are initially dispatched to each edge device from the cloud server by a Transmitter. Then, the edge device extracts its feature via a local visual backbone and compares this feature with all local gallery images, and the resulting similarity is used to create the uploading sequence (denoted as \textit{sequence} in Fig.~\ref{fig:comp}). Due to transmission limitations, there is an upper bound to the number of data the cloud server can accept at one time, so the uploading sequence of images is uploaded in batches. The user checks the sequence and terminates this process when its desired image is returned. Therefore, to achieve an efficient and effective inference, the user's desired image should be returned to the cloud server swiftly. 
Therefore, two key points in our framework are: (i) the edge device with the desired image should have a higher chance of uploading the image to the cloud server, and (ii) the desired image should be positioned at the beginning of the upload sequence.

To handle the points above, we specifically introduce a distribution-aware correlation modeling network (DaCM), which is deployed in both the cloud server to adjust the bandwidths of edge devices and each edge device to re-rank the image indexes in the uploading sequence. The input of DaCM is spatial-temporal data, \emph{i.e.} the timestamps and camera ID of images, which can be effortlessly obtained from the ReID system. 
Initially, it embeds spatial-temporal correlations into a graph structure by learning such a problem: what is the likelihood that an object will appear again in camera $j$ after a time delay $t$, from where it was previously observed in camera $i$. After training, the topology of the camera network and the movement rules of the object in the current scene will be embedded into DaCM implicitly, so as to support the adjustment of the bandwidth allocated to the edge devices and the index of the image. 

Furthermore, since we focus on a new ReID inference pattern, traditional evaluation protocols do not fully showcase the capabilities of proposed method. Thus, we propose several new evaluation protocols and their details will be described later. Finally, extensive experimental results demonstrate that our method improve the performance with a significant enhancement in accuracy and efficiency.

The contributions of our work can be summarised as:
\begin{itemize}
    \item To handle the rapidly growing number of videos, we propose an inference framework for ReID systems, which can evolve current methods into a cloud-edge collaborative pattern, enhancing both efficiency and effectiveness.
    \item To boost the system's performance by increasing the return probability of the desired image, we design a Distribution-aware Correlation Modeling network that captures the spatial-temporal correlations of the scene.
    \item  To demonstrate the superiority of our method, we introduce several evaluation protocols and conduct extensive experiments, with the results showcasing the significant enhancement achieved by our proposed framework.
\end{itemize}

\section{Related Work}
\label{sec:relatedwork}
\subsection{Object Re-identification}
Earlier ReID methods are type-specific, relying on specific attributes of the object, and are applicable only to a particular type of objects, such as person ReID~\cite{DBLP:conf/iccv/ZhengSTWWT15, DBLP:conf/cvpr/AhmedJM15, DBLP:conf/cvpr/ChengGZWZ16, DBLP:conf/iccv/ZhengZY17} and vehicle ReID~\cite{DBLP:conf/icmcs/LiuLMF16, DBLP:conf/eccv/LiuLMM16, DBLP:conf/mm/LiuLMC17}. 
As methods continue to advance, there is a growing trend towards developing generic ReID methods~\cite{DBLP:conf/cvpr/0004GLL019,DBLP:conf/cvpr/Huynh21,DBLP:conf/iccv/He0WW0021,DBLP:conf/aaai/LiSL23, DBLP:journals/pami/YeSLXSH22, DBLP:conf/cvpr/ChengGZWZ16,DBLP:conf/cvpr/SunCZZZWW20, DBLP:conf/iccv/He0WW0021} that are agnostic to the type of object being applied. 
% The focus is on creating general deep learning technologies, including feature extractors~\cite{DBLP:conf/cvpr/0004GLL019,DBLP:conf/cvpr/Huynh21,DBLP:conf/iccv/He0WW0021,DBLP:conf/aaai/LiSL23}, metric strategies~\cite{DBLP:journals/pami/YeSLXSH22}, and loss functions~\cite{DBLP:conf/cvpr/ChengGZWZ16,DBLP:conf/cvpr/SunCZZZWW20}. Recently, some methods~\cite{DBLP:conf/iccv/He0WW0021} adopts the Vision Transformer~\cite{DBLP:conf/iclr/DosovitskiyB0WZ21} as the backbone and achieves superior performance for both person and vehicle ReID tasks. 
All the above methods can be employed in our cloud-edge collaborative framework, partnering with DaCM for efficient and effective inference.

Since spatial-temporal information can be effortlessly obtained in a ReID system, some ReID methods~\cite{DBLP:conf/mmm/HuangHLYWZZ16, DBLP:journals/cviu/ChoKPLY19, DBLP:conf/aaai/WangLHX19} incorporate it to filter out unreasonable samples. Compared with them, our approach has several obvious differences : (i) Previous methods generate the spatial-temporal distribution through frequency statistics, whereas our approach employs a deep neural network to learn such correlations; (ii) Previous methods still adhere to centralized patterns, whereas our approach is developed within a cloud-edge collaborative framework; (iii) Previous methods only use such information to enhance performance, whereas our approach improves performance while achieving efficient inference. As a similar work, \citeauthor{DBLP:conf/ieeesec/JainZZAJSBG20} also interpolate such information in object searching, but the proposed Spatula directly filters out many candidate images that leads that (i) the desired image may not be found even with the replay strategy, and (ii) it is hard to combined with neural networks for promising performance.

\subsection{Cloud-Edge Collaborative Methods}
Emerging cloud-edge collaboration approaches showcase their superiority in various systems and communication technologies. Noteworthy instances of these advanced methodologies are evident in seminal works, such as the collaborative occluded face recognition architecture~\cite{DBLP:journals/tmm/ZhangHWYYT23}, the open-source framework SmartEye for real-time video analytic~\cite{DBLP:conf/mm/WangG21}, and the video service enhancement within an edge-cloud collaboration framework~\cite{DBLP:journals/tmm/WuBLWZW21}. The adaptation of cloud-device collaboration sensitive to changing environments~\cite{DBLP:conf/cvpr/GanPZLZLZ23}, the real-time surveillance video analysis in Cloud-Edge architecture~\cite{DBLP:conf/ijcnn/HouZ21}, and the Classification Driven Compression framework for reducing deep learning bandwidth consumption~\cite{DBLP:conf/ijcai/Dong0YZY20} further underscore the versatility and impact of these collaborative approaches. Unlike these methods, we focus on the ReID task and aim to achieve efficient and effective inference instead of model optimization. 

\section{Problem Definition}
\label{sec:cec-reid}
\begin{figure}[t]
    \centering
        \includegraphics[width=\linewidth]{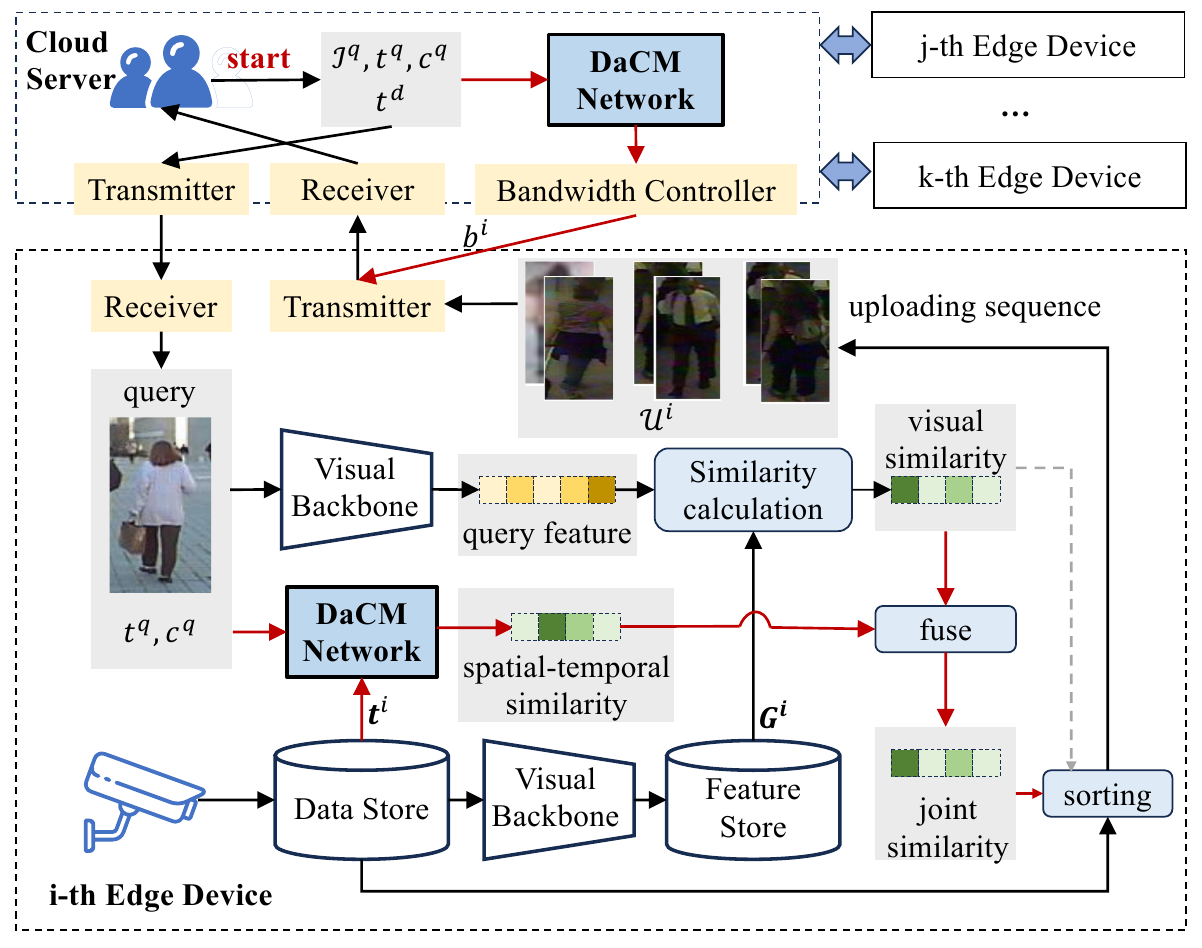}
        \caption{The overview of our proposed cloud-edge collaborative inference framework. The DaCM is deployed in both the cloud server and edge devices for adjusting uploading batch size $b^i$ and image order in the uploading sequence. The red solid denotes the data flow enabled by the designed DaCM, and the gray dashed line denotes the previous data flow that can be removed by DaCM.}~\label{fig:overview}
    \end{figure}

As shown in Fig.~\ref{fig:overview}, given a query image $\mathcal{I}^{q}$ and auxiliary information $\{t^{q}, c^{q}, t^{d}\}$ ($t^{q}$ and $c^{q}$ denote the timestamp and camera ID of $\mathcal{I}^{q}$, respectively, while $t^{d}$ represents the target time of the desired image), the ReID system sends these information to each edge device. Then, for the $i$-th edge device, it extracts the deep feature $\mathbf{f}^q$ from $\mathcal{I}^{q}$ via a local visual backbone and compute the similarity $\mathbf{s}^i\in\mathbb{R}^{N_i}$ between $\mathbf{f}^q$ and the features $\mathbf{G}^i \in \mathbb{R}^{N_i\times E}$ of all $N_i$ gallery images on the $i$-th device ($E$ is the dimension). The images in current edge device are ranked by $\mathbf{s}^i$ and sent to the cloud server in batches due to the limited bandwidth. Finally, users check the returned data, and the process can be terminated if the \textit{desired images} are contained in current batch.

Denoting the user desired image as $\mathcal{I}^d$, we can see that the performance of cloud-edge collaborative ReID systems depends on when the $\mathcal{I}^d$ is returned to the server, which is influenced by: (1) the delay of the communication network; (2) the speed of feature extraction; (3) the rank of $\mathcal{I}^d$ in the uploading sequence, and (4) the amount of data that the camera (on which $\mathcal{I}^d$ is captured) can upload each time, i.e. the bandwidth allocated for the camera. The first two problems have been well studied by previous methods~\cite{DBLP:journals/corr/abs-2008-11560,DBLP:conf/mm/Zhuang0Z21,DBLP:conf/ieeesec/JainZZAJSBG20,10.14778/3137628.3137664,DBLP:conf/mobicom/ZhangZLMHL0MC20}, but the last two problems still lack relevant methods. Therefore, in this paper, we focus on how to accelerate the system by advancing the position of $\mathcal{I}^d$ in the sequence and increasing the bandwidth utilization rate of the camera where $\mathcal{I}^d$ is located from the perspective of multimedia computing. The optimization objective to reduce the Transmission Number (TN) can be expressed as:
\begin{equation}~\label{eq:tn}
    \begin{aligned}
         \arg\min_{\theta} \left(\Omega_i \left(\left\lceil\frac{\epsilon(\mathcal{U}^i(\mathbf{s}^i))}{b^i}\right\rceil \right)\right),
    \end{aligned}
\end{equation}
where $\mathcal{U}^i$ is the uploading image sequence of $i$-th edge device, which is determined by the score $\mathbf{s}$, $b^i$ is the allocated bandwidth for $i$-th edge device, and it means how many images can be uploaded each time, $\theta$ is the parameters should be optimized, which influence $\mathcal{U}^i$ and $b^i$.
Function $\epsilon$ is used to return the index of $\mathcal{I}^d$ in $\mathcal{U}^i$, function $\Omega$ aggregate the results from all edge devices, and their implementations are contingent upon user requirements. 

% For example, in the conventional ReID task, it is acceptable for the system to just return \textit{one} image with ID $l^q$, so $\epsilon$ represents the index of the first image in $\mathcal{U}^i$ with ID $l^q$, and $\Omega$ can be implemented by a minimum function. If $\mathcal{I}^d$ is not contained in $i$-th edge device, $\epsilon$ should output positive infinity, and we use a much larger value in experiments. In this paper, we use $t^d$ to specify the desired image.

Most of previous methods pay much attention to learning a proper $\mathcal{U}^i$, i.e. forcing the images with same identity ID $l^q$ have small $\mathbf{s}$ to make them at the front of the sequence, and they can not have an impact on $b^i$. In contrast, we propose the DaCM network, which can learn the spatial-temporal distribution of objects in scene and be used to boost the efficiency of the system by adjusting both $\mathcal{U}^i$ and $b^i$.

\section{Proposed Approach}
\label{sec:streid}
\begin{figure}[t]
    \centering
        \includegraphics[width=0.8\linewidth]{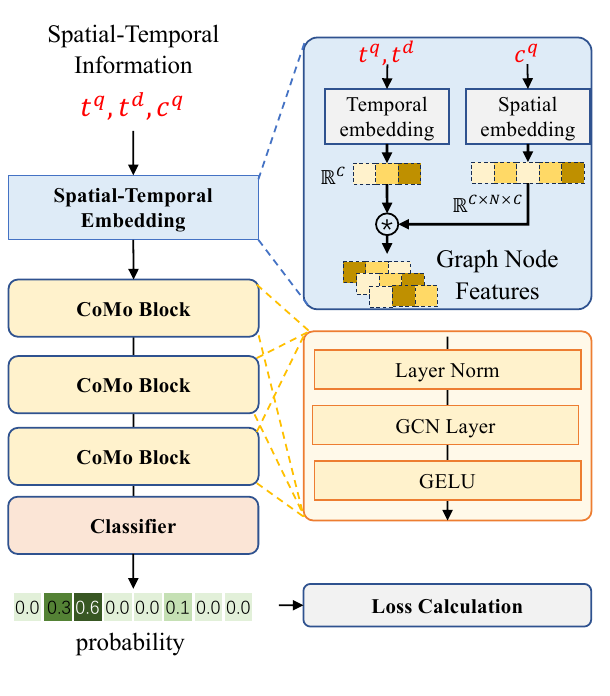}
        \caption{The architecture of DaCM network.}~\label{fig:network}
\end{figure}

In this section, we first present the details of DaCM architecture and its training strategy, then describe how DaCM is used in the cloud-edge collaboration framework.
\subsection{DaCM Architecture}
The DaCM network performs an important role in the cloud-edge collaborative ReID system, and in this part, we describe its architecture. As shown in Fig.~\ref{fig:network}, DaCM mainly consists of three components, a spatial-temporal embedding module, multiple Correlation Modeling (CoMo) blocks, and a final classifier.

\noindent\textbf{Spatial-temporal embedding.} 
Inspired by the positional encoding manner used in~\cite{DBLP:conf/nips/VaswaniSPUJGKP17}, we first adopt the sinusoidal embedding to encode the temporal information. Denoted the timestamps of query image and target as $t^q$ and $t^d$, we encode their difference to a feature vector in the formulation of:
\begin{equation}
\label{eq:tem-emb}
    \begin{aligned}
        \mathbf{e}_{2i}   = \sin\left(\frac{(t^d - t^q)}{\lambda^{\frac{2i}{D}}}\right), ~  
        \mathbf{e}_{2i+1} = \cos\left(\frac{(t^d - t^q)}{\lambda^{\frac{2i}{D}}}\right),
    \end{aligned}
\end{equation}
where $i$ is the dimension index, $\mathbf{e}\in\mathbb{R}^{D}$ is the results embedding, and $D$ is the dimension of the embedding. $\lambda$ denotes the max period of the sinusoidal function. The wavelengths form a geometric progression from $2\pi$ to $\lambda\cdot2\pi$. 

For the spatial information, considering the fixed topology of different cameras, we suggest employing learnable parameters $\mathbf{W}\in\mathbb{R}^{C\times D \times C\times D}$ to represent the spatial information. $C$ is the number of cameras and the nodes in the graph. When $c^q$ is provided as input, we choose $\mathbf{W}_{c^q}$ as the resulting spatial embedding, which is then combined with $\mathbf{e}$ to generate a graph structure:
\begin{equation}
    \begin{aligned}
        \mathbf{A} = \frac{1}{\sum_j^D e_j} \sum_j^D \mathbf{W}_{c^q, j} * e_j + \mathbf{b} \in \mathbb{R}^{C\times D},
    \end{aligned}
\end{equation}
where $\mathbf{b}$ is a learnable bias.

\noindent\textbf{Correlation modeling block.} As for CoMo blocks, they are used to propagate the information among graph nodes, and as shown in Fig.~\ref{fig:network}, one CoMo block is comprised of a Layer Normalization, a GCN layer~\cite{DBLP:conf/iclr/KipfW17}, and a GeLU activation, which can be formulated as:
\begin{equation}
    \begin{aligned}
        \mathbf{A}^{l+1} =\operatorname{GELU}\left( \mathbf{E}^l \times \operatorname{LN}(\mathbf{A}^{l}) \times \mathbf{W}^l  \right),
    \end{aligned}
\end{equation}
where $\mathbf{E}^l \in \mathbb{R}^{C\times C}$ is the adjacency matrix of the GCN layer, which can be learned during training, and $\mathbf{W}^l \in \mathbb{R}^{C\times C}$ denotes the transfer weight to update the information for each node feature. 

\noindent\textbf{Classifier.} Finally, we apply an MLP for each node feature in the graph as the classifier, whose output dimension is set to 1. The outputs of all nodes are concatenated as the final output of the DaCM network:
\begin{equation}
    \label{eq:gnnoutput}
    \begin{aligned}
        \mathbf{y} = [\operatorname{MLP}(\mathbf{G}_0); \ldots; \operatorname{MLP}(\mathbf{G}_{C-1})] \in \mathbb{R}^{C}.
    \end{aligned}
\end{equation}
The MLP consists of the sequence of BatchNorm~\cite{DBLP:conf/icml/IoffeS15}, ReLU, and fully connected layers. 

\subsection{Objective Function}
In this part, we describe how to train the DaCM network. For the training set $\mathcal{D}^\mathrm{train}$, we randomly select two samples ($q$ and $d$) with the same label ID but different camera IDs, and their spatial-temporal information is denoted $\{l^q, t^q, c^q\}$ and $\{l^d, t^d, c^d\}$, respectively. According to the time difference $t^d - t^q$ and camera id $c^q$, we obtain the output $\mathbf{y}$ via the proposed network, and then we optimize the network via the expectation: 
\begin{equation}
    \begin{aligned}
        \arg\max_\theta\mathbb{E}_{(q, d) \sim \mathcal{D}^\mathrm{train}}[P(y=c^d| t^q, t^d, c^q; \theta)],
    \end{aligned}
\end{equation}
where $\theta$ is the parameters of the network. In implementation, this expectation can be easily converted to a classification problem, and we adopt the cross-entropy loss function to generate the gradients:
\begin{equation}
    \begin{aligned}
        \ell = - \log \frac{\exp(\mathbf{y}_{c^d})}{\sum_i^C \exp(\mathbf{y}_{i})},
    \end{aligned}
\end{equation}
where $y$ is the output of the DaCM network. The classification solves such a problem: determining the camera at which an object will appear again after a duration of $t^d - t^q$, starting from camera $c^q$. the network will acquire knowledge about the topology of devices deployed in the system and some characteristics of the target's movement, thereby facilitating efficient inference.

\subsection{Inference}
As discussed in Sec.~\ref{sec:cec-reid}, to achieve efficient inference, the system should promptly return the desired image to the cloud server by generating appropriate $\mathbf{s}$ and $b$. In this section, we elaborate on how to accomplish this goal by using the DaCM network.

\noindent\textbf{Cloud-level inference.}
For cloud-level inference, its emphasis is on assigning a large $b^i$ to the edge devices containing the desired image. Assuming the total bandwidth allocated for $C$ edge devices is $B$, a simple strategy would be to distribute $B$ equally among edge devices, but it lacks flexibility. We aim to decrease data transmission in the connected network and alleviate stress on the system by dynamically allocating bandwidth to the edge devices based on the spatial-temporal correlation learned in the DaCM network.

Given the query information of $\{\mathcal{I}^q, t^q, c^q, t^d\}$, we send the spatial-temporal information into the DaCM network, and it produces $\mathbf{y} \in \mathbb{R}^{C}$. The output representation is the chance of the target appearing under each camera at moment $t^d$. Intuitively, if $\mathbf{y}_i > \mathbf{y}_j$, the $i$-th edge device should be allocated with a larger bandwidth than the $j$-th edge device. Thus, we formulate this process as:
\begin{equation}
    \begin{aligned}
        \hat{b}^i = \operatorname{softmax}(\mathbf{y}/\gamma_0)_i * B,
    \end{aligned}
\end{equation}
where $\gamma_0$ is used to smooth the probability to avoid extreme values. However, it overlooks a crucial factor—the uneven distribution of data among edge devices. The number of images on the edge devices can vary significantly, necessitating the consideration of this factor when allocating bandwidth. Therefore, we finally assign the bandwidth $b^i$ for the $i$-th edge device as:
\begin{equation}
\label{eq:band}
    \begin{aligned}
        b^i = \frac{z^i}{\sum_j z^j} * B,~ z^i = \phi\left(\frac{\mathbf{y}}{\gamma_0}\right)_i * \left(\frac{\exp(|\mathcal{G}^i|)}{\gamma_1\sum_j\exp(|\mathcal{G}^j|)}\right),
    \end{aligned}
\end{equation}
where $\phi$ is the $\operatorname{softmax}$ function. Note that for cloud-level inference, $t^d$ should be provided by the user.

\noindent\textbf{Edge-level inference.}
For edge-level inference, the focus is on re-ranking the index of gallery images in the uploading sequence by generating proper $\mathbf{s}^i$. Denoted the query image with its spatial-temporal information as $\{\mathcal{I}^q, c^q, t^q\}$, one gallery image at the $c^d$-th edge device as $\{\mathcal{I}^d, c^d, t^d\}$, we send $\{c^q, t^q, t^d\}$ into DaCM and obtain the output $\mathbf{y}$. If $\mathbf{y}_{c^q}$ is small, it implies that $\mathcal{I}^q$ and $\mathcal{I}^d$ do not match in spatial-temporal correlation, resulting in a minimal likelihood of having the same ID as the query image. Applying this operation to all gallery images on the $i$-th edge device, we obtain spatial-temporal similarity. Next, we delve into how to combine such spatial-temporal similarity with visual similarity.

Given a reliable visual similarity, it is difficult to build a reliable joint metric because the spatial-temporal similarity is unreliable and it is hard to assign appropriate weighting factors for these two types of metrics. Inspired by the joint metric proposed in~\cite{DBLP:conf/aaai/WangLHX19}, we adopt a smoothing operator to alleviate unreliable probability estimation. Denoted the spatial-temporal similarity as $\mathbf{o}\in\mathbb{R}^{N_i}$ and the visual similarity as $\mathbf{v}\in\mathbb{R}^{N_i}$ (assume it is produced by cosine distance function, and large value in $\mathbf{v}$ means the two features are similar), where $N_i$ is the number of gallery images in the $i$-th device, the joint similarity is computed as:
\begin{equation}
\label{eq:joint_sim}
    \begin{aligned}
        s_{k}^i = - \frac{1}{1+\alpha\exp\left(\phi(- \frac{\mathbf{o}}{\beta})_k\right)} \frac{1}{1+\exp(\mathbf{v}_k-1)},
    \end{aligned}
\end{equation}
where $\alpha$ and $\beta$ are two hyper-parameters to balance these two similarities, and the gallery images are re-ranked according to $\mathbf{s}^i$.

\noindent\textbf{Time-constrained ReID.}
For a ReID system, users sometimes wish to search for targets near a specific time, a task challenging to accomplish solely based on visual features. Sorting only by time may introduce a large number of unrelated images. Therefore, the key to achieving tcReID lies in how to effectively combine time information with visual information. We observe that Eq.~\eqref{eq:joint_sim} offers a natural way to fulfill such a task. However, Eq.~\eqref{eq:joint_sim} does not satisfy the requirement because it does not introduce $t^d$ into the similarity calculation. Therefore, we propose a new formulation to meet the requirements of tcReID task. 

Denoted the query data and target as $\{\mathcal{I}^q, t^q, c^q, t^d\}$, we construct a pattern bank by calculating the correlation between the query image and gallery images in the edge device. DaCM takes in $\{\mathcal{I}^q, t^q, c^q, t^{g_i}\}$ ($t^{g_i}$ is the timestamp of $g_i$-th gallery image) and output $\mathbf{a}^{g_i} \in\mathbb{R}^C$. This process is applied to all gallery images and we collect them as a pattern bank $\mathbf{B}\in\mathbb{R}^{N_i\times C}$ of $\mathcal{I}^q$. Then we send the true target time and query data into DaCM and output $\mathbf{y}$. We calculate the similarity for constructing an uploading image sequence in the form of:
\begin{equation}
\label{eq:tcreid-sim}
    \begin{aligned}
        \hat{s}_{k}^i =  \frac{s_k^i}{1+\exp\left(\cos(\mathbf{B}_k, \mathbf{y})-1\right)},
    \end{aligned}
\end{equation}
where $\cos$ denotes the cosine distance function. Finally, we sort the gallery images according to $\hat{\mathbf{s}}^i$ and return them to the cloud server in batches. In addition, we find that the solely using the outputs of DaCM may lead to outlier problems. Therefore, in order to ensure the stability, we combine the output of DaCM with the spatial-temporal correlation obtained by the frequency statistics method~\cite{DBLP:conf/aaai/WangLHX19}.

\section{Experiments}
\label{sec:exp}
\subsection{Experimental Settings}\label{sec:5.1}
\noindent{\bf Datasets.} We mainly evaluate our proposed framework and method on the DukeMTMC-reID~\cite{DBLP:conf/iccv/ZhengZY17} and Market-1501~\cite{DBLP:conf/iccv/ZhengSTWWT15} datasets, since they are annotated with high-quality timestamp.

\noindent{\bf Compared methods.}
We compare our method with several inference strategies to show its performance, including:
\begin{itemize}
    \item \textit{Pattern-C} denotes the conventional centralized inference strategy, which collects all images captured by edge devices and conducts similarities calculations in the cloud server. We use it as the baseline to show the boosting effectiveness of different inference strategies.
    \item \textit{Pattern-CE} denotes a simple cloud-edge collaborative inference strategy: the amount of transmission is evenly distributed to each edge device and each edge device assigns the upload sequence according to the distance between the query image and gallery images. By comparing with this strategy, we can see the improvements brought by the proposed DaCM network.
\end{itemize}
Besides,  we design a DaCM network to boost the efficiency of the ReID system, and we also replace the DaCM in the system with \textit{stReID}~\cite{DBLP:conf/aaai/WangLHX19}, which uses frequency statistics to model spatial-temporal associations, and it interpolates the statistical spatial-temporal distribution into the similarity calculation process for person ReID. 

\noindent\textbf{Hyper-parameters:} To train the DaCM network, we employ Adam~\cite{DBLP:journals/corr/KingmaB14} as the optimizer. The initial learning rate is set to 0.01 and is reduced by 10 for every 30 epochs. $\gamma_0$ and $\gamma_1$ are both set to 0.01 as the default. $\alpha$ and $\beta$ are bot set to 0.1. $\lambda$ in Eq.~\eqref{eq:tem-emb} is set to 10,000 as the default. $B$ is set to $3*C$, i.e., each edge device can upload an average of three images at a time.

\subsection{Evaluation Protocols}
We propose several novel protocols to show the performance of the cloud-edge collaborative inference. Let us initially provide a definition of the \textit{desired image}, as the proposed protocols hinge upon this conceptual foundation. A desired image is a particular sample among the gallery images, sharing the same identity as the query image and possessing a timestamp in proximity to a given target time. 

\begin{itemize}
    \item \textit{mean Transmission Number(mTN)}: it is a protocol used to present the efficiency of the method. For each pair of one query image and one gallery image with same identity ID, there exists one corresponding TN as shown in Eq.\eqref{eq:tn}, and we average the TN of all pairs as mTN.
    
    \item \textit{precise Rank@K (PR-K)}: it is calculated by checking whether top-k gallery images contain the desired image that has the same ID with the query image and is closest to the target time, so pR-K is a stricter protocol than R-K.
    
    \item \textit{mean precise Rank (mpR)}: For the $i$-th query data, when pR-$k_i$ is successful but pR-($k_i$-1) is not successful, we record its precise Rank as $k_i$, and we average the $k_i$ of all query data as mpR.
\end{itemize}

\subsection{Experimental Results}

\noindent\textbf{Boosting effect of DaCM for ReID methods.}
Since the edge-level inference in our method can be seen as one kind of re-ranking technologies, we embedded it to several visual ReID methods (PCB~\cite{DBLP:conf/eccv/SunZYTW18}, SBS~\cite{DBLP:conf/mm/HeLLLCM23}, TransReID~\cite{DBLP:conf/iccv/He0WW0021}) to show the boosting performance. 
Methodologies for comparison can be categorized into several different groups, including several classical methods such as LOMO+XQDA~\cite{DBLP:conf/cvpr/LiaoHZL15} and handcrafted approach BoW+kissme~\cite{DBLP:conf/iccv/ZhengSTWWT15}, explicit deep learning methods including PAN~\cite{DBLP:journals/tcsv/ZhengZY19}, SVDNet~\cite{DBLP:conf/iccv/SunZDW17} and HA-CNN~\cite{DBLP:conf/cvpr/LiZG18}, attribute-centric techniques including APR~\cite{DBLP:journals/pr/LinZZWHYY19}, mask-guided strategies including Human Parsing~\cite{DBLP:conf/cvpr/KalayehBGKS18}, part-based approaches like PSE+ECN~\cite{DBLP:conf/cvpr/SarfrazSES18}, pose-oriented techniques like PCB~\cite{DBLP:conf/eccv/SunZYTW18}, and a recent work CLIP-ReID~\cite{DBLP:conf/aaai/LiSL23}.

The results evaluated on DukeMTMC-reID dataset for comparison are shown in Table~\ref{tab:boosting-duke}. Without bells and whistles, our method outperforms all existing methods on the DukeMTMC-reID dataset. 
In addition, the robustness of our methodology is further highlighted when employing the same visual stream method. For instance, integrated with SBS~\cite{DBLP:conf/mm/HeLLLCM23}, our approach outperforms st-ReID~\cite{DBLP:conf/aaai/WangLHX19}, elevating the rank-1 accuracy from 95.4\% to 96.7\%, and boosting mAP from 83.0\% to 89.8\%. Besides, the results evaluated on Market1501 dataset for comparison are shown in Table~\ref{tab:boosting-market}, and our method still gain obvious improvements than the baselines in term of the R-1.

\begin{table}[t]
\centering
\caption{Boosting Effect on DukeMTMC-reID dataset. }\label{tab:boosting-duke}
    \begin{tabular}{l|c|c}
        \hline
        Methods & R-1$\uparrow$    & mAP$\uparrow$ \\ \hline
        BoW+kissme~\cite{DBLP:conf/iccv/ZhengSTWWT15} & 25.1   & 12.2  \\
        LOMO+XQDA~\cite{DBLP:conf/cvpr/LiaoHZL15} & 30.8   & 17.0  \\
        PAN~\cite{DBLP:journals/tcsv/ZhengZY19}   & 71.6   & 51.5  \\
        SVDNet~\cite{DBLP:conf/iccv/SunZDW17} & 76.7   & 56.8  \\
        HA-CNN~\cite{DBLP:conf/cvpr/LiZG18} & 80.5    & 63.8  \\
        APR~\cite{DBLP:journals/pr/LinZZWHYY19}   & 70.7  & 51.9  \\
        Human Parsing~\cite{DBLP:conf/cvpr/KalayehBGKS18} & 84.4   & 71.0  \\
        PSE+ECN~\cite{DBLP:conf/cvpr/SarfrazSES18} & 85.2   & 79.8  \\
        % MultiScale~\cite{DBLP:conf/iccvw/ChenZG17} & 79.2    & 60.6  \\
        CLIP-ReID~\cite{DBLP:conf/aaai/LiSL23} & 90.0     & 80.7  \\
        \hline 
        PCB~\cite{DBLP:conf/eccv/SunZYTW18}   & 82.3    & 70.7  \\
        PCB + stReID~\cite{DBLP:conf/aaai/WangLHX19} & 94.3   & 84.0 \\ 
        PCB + InSTD~\cite{DBLP:conf/iccv/RenHLLWT21}   & 92.7  & 86.1  \\ 
        PCB + Ours (OE) & \textbf{96.2}  & \textbf{ 89.5}  \\ 
        \hline 
        SBS~\cite{DBLP:conf/mm/HeLLLCM23}   & 90.8   &  79.9           \\
        SBS + stReID & 95.4    & 83.0  \\
        SBS + Ours (OE)& \textbf{96.7}   & \textbf{89.8}  \\ 
        \hline 
        TranReID~\cite{DBLP:conf/iccv/He0WW0021} & 90.8    & 81.8  \\ 
        TranReID + stReID &  96.2 &  88.6 \\
        TranReID + Ours (OE)& \textbf{96.8}   & \textbf{91.0}  \\ 
        \hline 
    \end{tabular}
\end{table}

\begin{table}[t]
\centering
\caption{Boosting Effect on Market-1501 dataset.}\label{tab:boosting-market}
    \begin{tabular}{l|c|c}
        \hline
        Methods & R-1$\uparrow$   & mAP$\uparrow$ \\ \hline
        BoW+kissme~\cite{DBLP:conf/iccv/ZhengSTWWT15} & 44.4    & 20.8  \\
        PAN~\cite{DBLP:journals/tcsv/ZhengZY19}   & 82.8    & 63.4  \\
        SVDNet~\cite{DBLP:conf/iccv/SunZDW17} & 82.3    & 62.1  \\
        HA-CNN~\cite{DBLP:conf/cvpr/LiZG18} & 91.2    & 75.7  \\
        APR~\cite{DBLP:journals/pr/LinZZWHYY19}   & 84.3    & 64.7  \\
        Human Parsing~\cite{DBLP:conf/cvpr/KalayehBGKS18} & 93.9   & -  \\
        PSE+ECN~\cite{DBLP:conf/cvpr/SarfrazSES18} & 90.3   & 84.0  \\
        CLIP-ReID~\cite{DBLP:conf/aaai/LiSL23} & 95.7   & 89.8  \\
        \hline 
        SBS~\cite{DBLP:conf/mm/HeLLLCM23}   & 95.8    & \textbf{89.0}            \\
        SBS + stReID & 96.1   & 86.8  \\
        SBS + Ours(OE)& \textbf{96.4}    & 88.2  \\ 
        \hline 
        TranReID~\cite{DBLP:conf/iccv/He0WW0021} & 95.2  & 89.0  \\ 
        TranReID + stReID & 96.6     & \textbf{89.8}  \\
        TranReID + Ours(OE)& \textbf{96.9}   & 89.0  \\ 
        \hline 
    \end{tabular}
\end{table}

\begin{table}[h]
    \caption{Performance of Different Inference Strategies Evaluated on DukeMTMC-reid Dataset. 
    }\label{tab:main-results2}
	\centering
	\renewcommand\arraystretch{1.1}
	\setlength{\tabcolsep}{1mm}{
        \begin{tabular}{l|c|c|c|c}
        \hline 
        Methods     & C & CE & OC & OC+OE \\ \hline
        mTN$\downarrow$ & 1561 & 9.56 & 5.39 & \textbf{4.43} \\
        \hline 
        \end{tabular}
    }
\end{table}

\begin{table}[h]
    \caption{Performance of Different Inference Strategies Evaluated on DukeMTMC-reid Dataset.
    }\label{tab:main-results1}
	\centering
	\renewcommand\arraystretch{1.1}
	\setlength{\tabcolsep}{1mm}{
        \begin{tabular}{l|c|c|c|c}
        \hline 
        protocol     & C  & Ours(Lin) & Ours(Sam) & Ours \\ \hline
        pR-1$\uparrow$ & 0.94  & 20.38 & 23.37 & \textbf{44.74}\\
        mpR$\downarrow$ & 123.77 & 20.56 & 9.35 & \textbf{1.48} \\ \hline
        \end{tabular}
    }
\end{table}

\noindent\textbf{Efficiency of the proposed ReID system.} We compare the proposed approach with the strategies introduced in Sec.~\ref{sec:5.1}, and the results are shown in Table~\ref{tab:main-results2} and Table~\ref{tab:main-results1}, where \textit{C} and \textit{CE} denotes the \textit{Pattern-C} and \textit{Pattern-CE}, respectively. \textit{OC} and \textit{OE} denotes using cloud-level inference and edge-level inference. By analyzing the mTN values of different strategies, we can see that \textit{Pattern-C} obtains a huge number of mTN since it requires uploading all images to the cloud server, and a simple cloud-edge collaborative framework (\textit{Pattern-CE}) reduces mTN to 9.56, which saves much network traffic. Meanwhile, the results in the table also show that using DaCM network alone in the cloud server or in the edge devices can reduce the mTN to a certain extent (from 9.56 to 5.39 and 4.43, respectively), and the combination of them can lead to an optimal result. As for the protocols of pR-1 and mpR, most of the previous methods do not take it into consideration, and their methods only produce meaningless output. As shown in Table~\ref{tab:main-results1}, if we use the visual similarity, it only achieves 0.94 pR-1. However, we can learn that our method can achieve 44.74 pR-1, which is still low but it demonstrates that the proposed approach can be applied to the challenging task. It is expected that more methods using spatial-temporal correlation will be proposed to solve this problem in the future.

\begin{figure}
	\centering
	\includegraphics[width=0.46\linewidth]{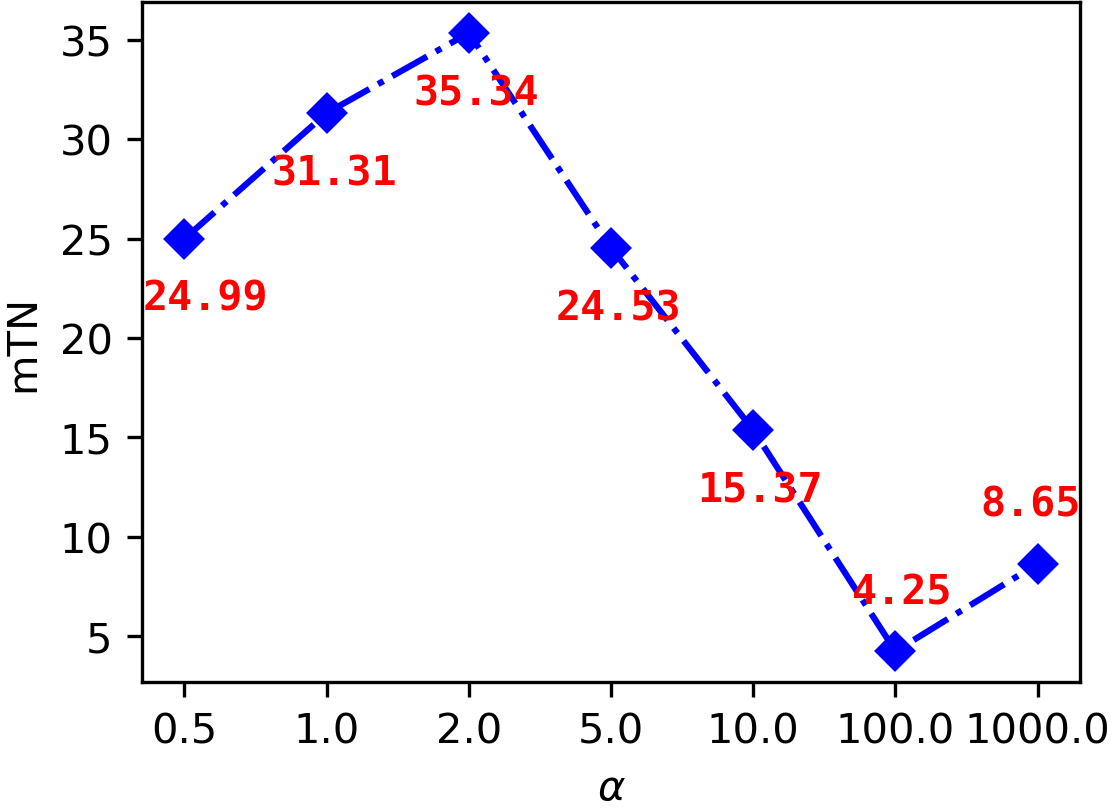}
	\includegraphics[width=0.46\linewidth]{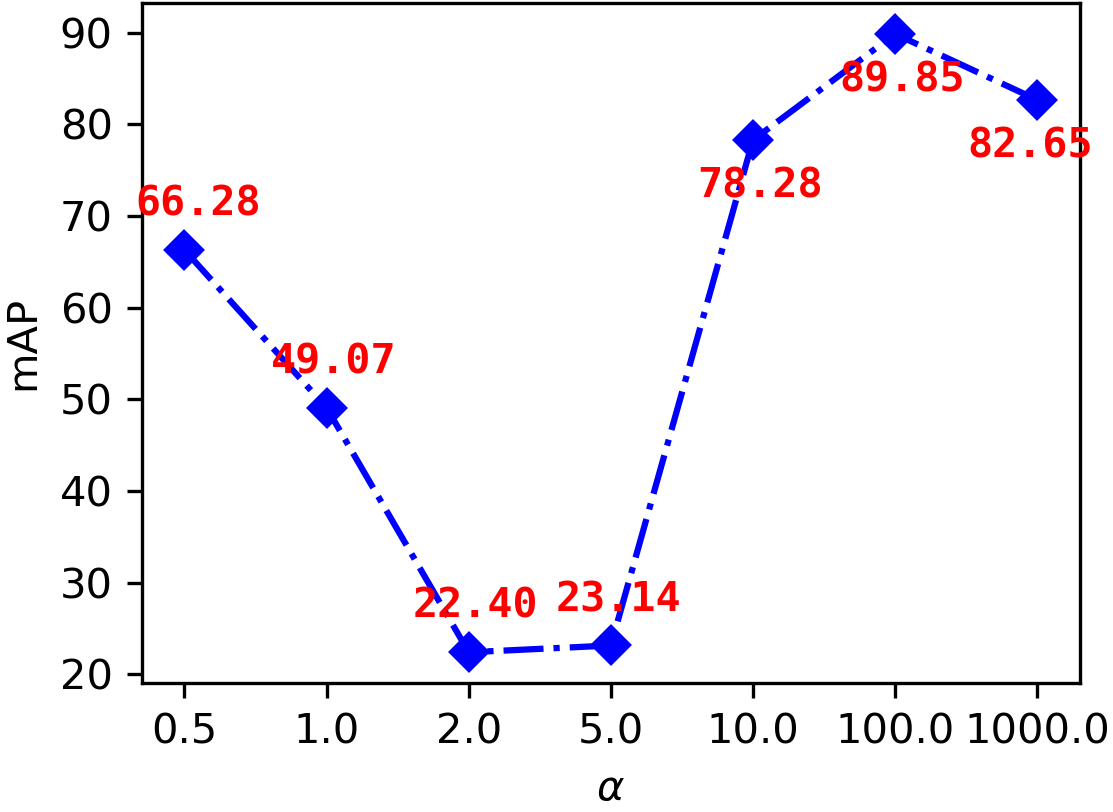}

	\includegraphics[width=0.46\linewidth]{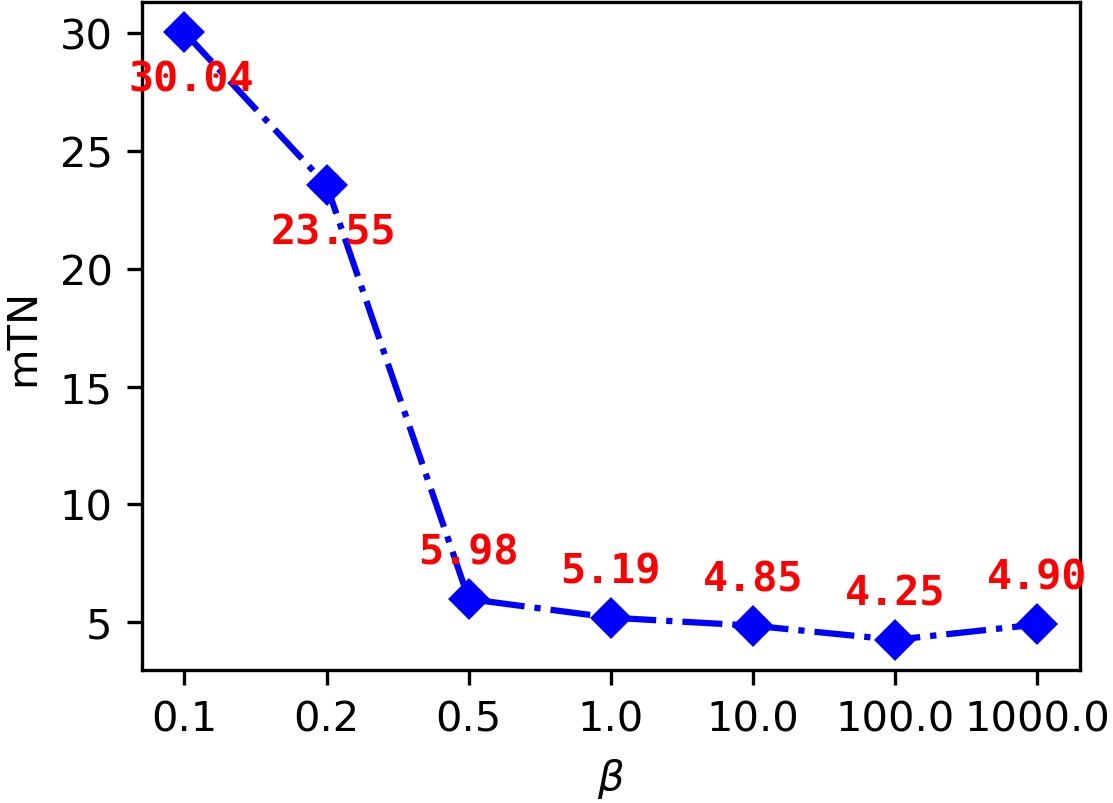}
	\includegraphics[width=0.46\linewidth]{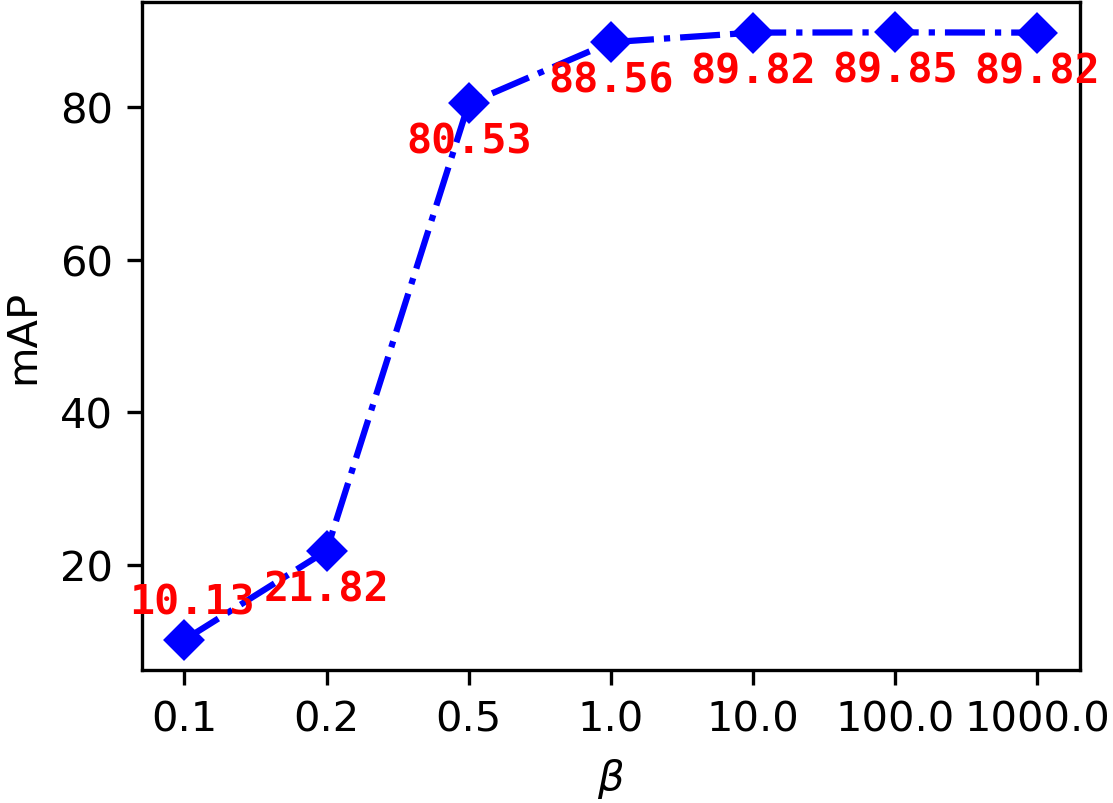}
	\caption{The effects of different values of $\alpha$ (upper) and $\beta$ (lower) for mTN and mAP.}
\label{fig:d_num}
\end{figure}

\begin{figure}
	\centering
	\includegraphics[width=0.47\linewidth]{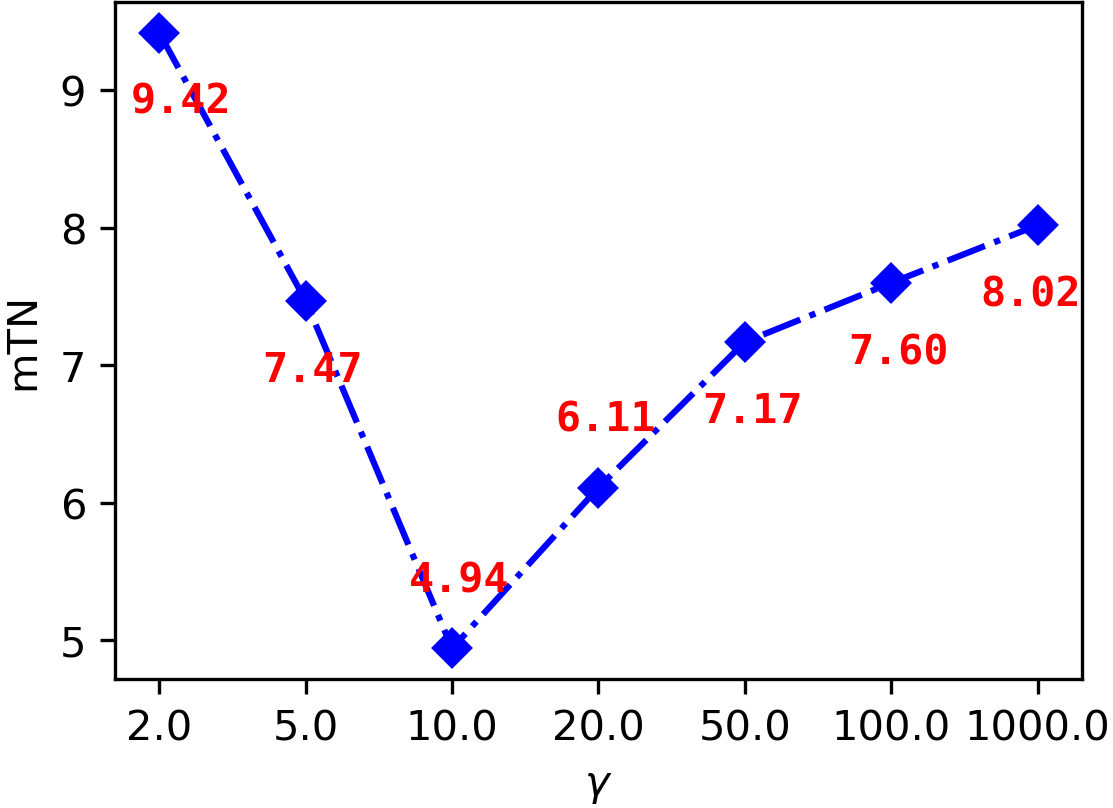}
	\includegraphics[width=0.47\linewidth]{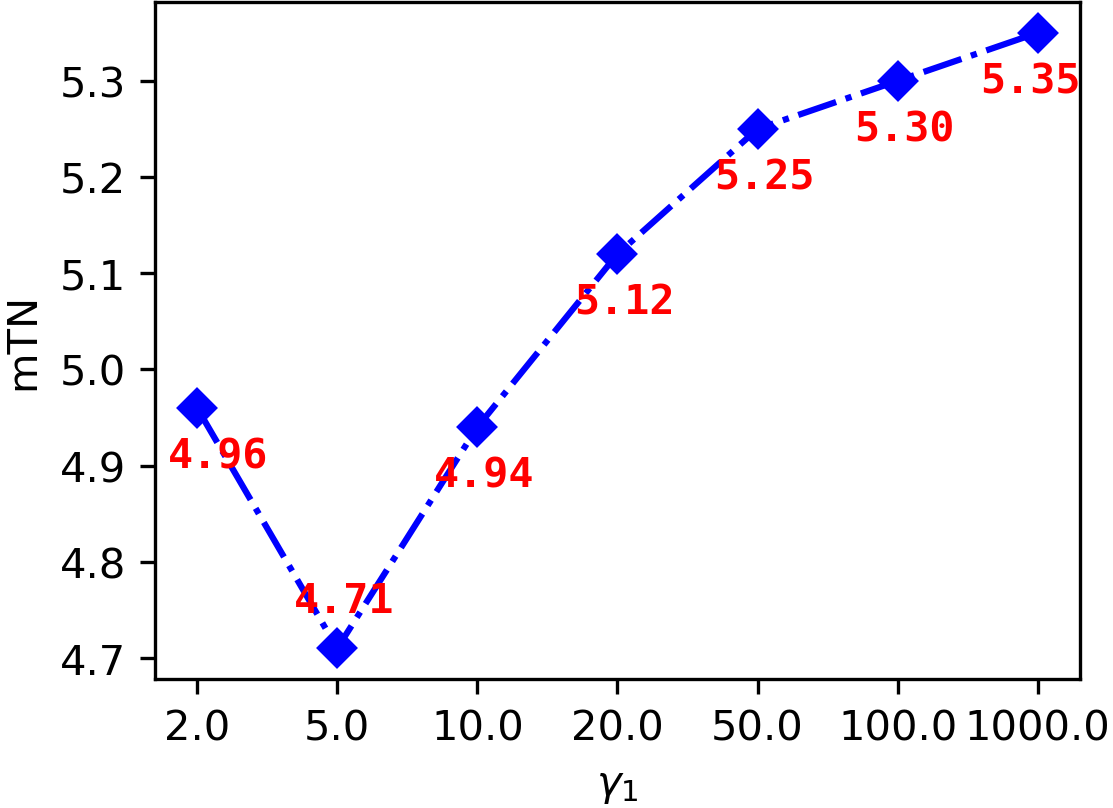}
	\caption{Effects of various $\gamma_0$ (left) and $\gamma_1$ (right) for mTN.}
\label{fig:g_num}
\end{figure}

\subsection{Ablation Study}

\noindent\textbf{Selection of different $\alpha$ and $\beta$.} $\alpha$ and $\beta$ are two hyper-parameters used in Eq.~\ref{eq:joint_sim}, which will affect the image order in the uploading sequence. Thus, we conduct two sensitivity analysis experiments to investigate their impact on our ReID system. For the protocol of mTN, we only show the results generated by only using the DaCM network in the edge devices. As shown in Fig.~\ref{fig:d_num}, when $\alpha$ is in the range of 1.0$\sim$10.0, the system has much worse performance, and when $\alpha$ is set to 100, the system achieves a low mTN value and a high mAP value. Besides, the performance improves as the value $\beta$ increases.

\noindent\textbf{Selection of different $\gamma_0$ and $\gamma_1$.} $\gamma_0$ and $\gamma_1$ are used in Eq.~\eqref{eq:band} to adjust the bandwidth assigned for each edge devices. Thus, we adjust their different values to show their impact on system performance. The results are shown in Fig.~\ref{fig:g_num}. Since these two hyper-parameters only affect the traffic of the connected network, we only present their effect for the protocol of mTN.
\begin{figure}[]
    \centering
    \includegraphics[width=\linewidth]{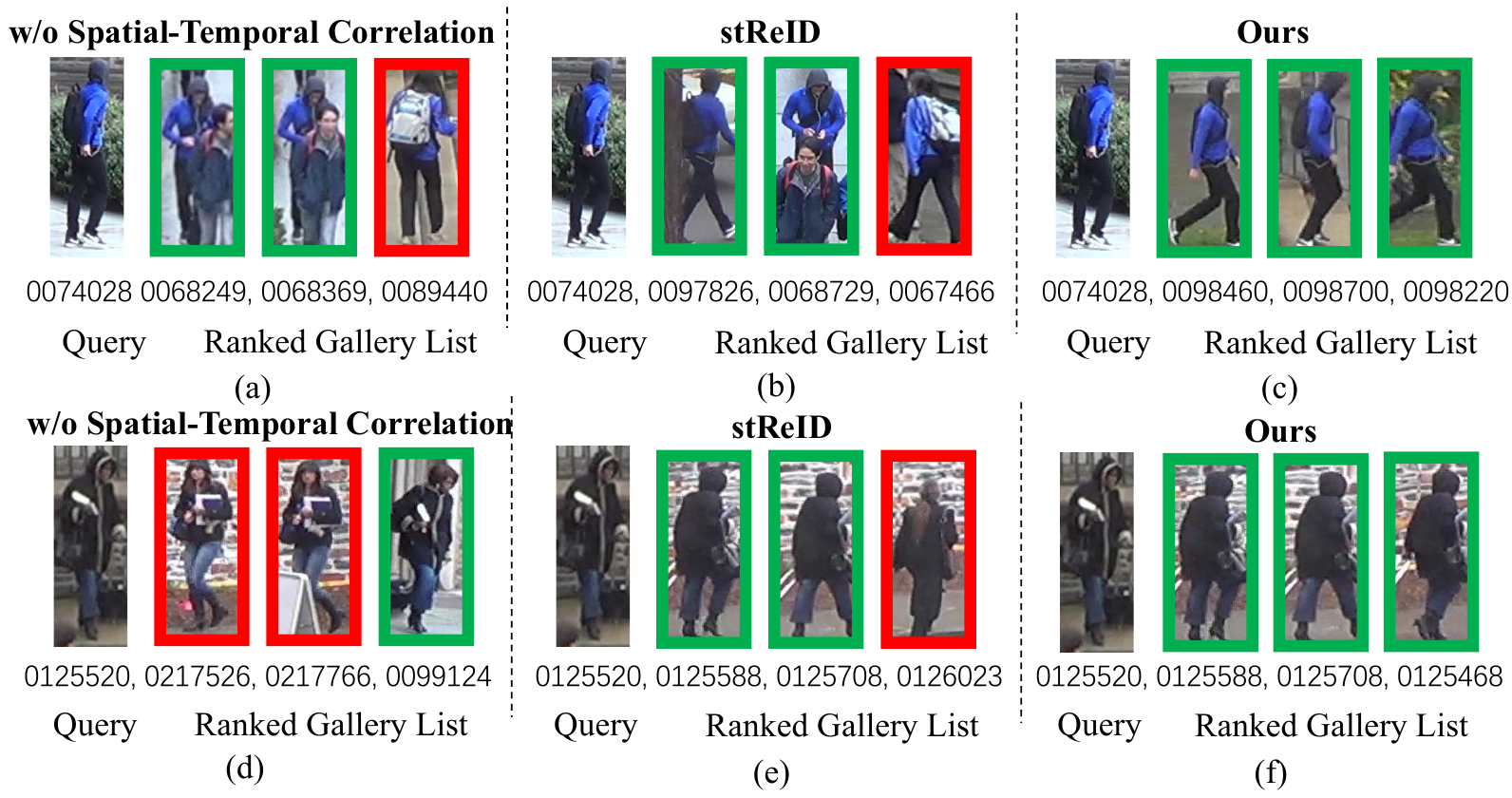}
    \caption{The visualization of retrieval results for some samples from the DukeMTMC-reID dataset. Timestamps (Frame ID) are shown below the images.}
\label{fig:vis}
\end{figure}

\subsection{Visualization}

To help understand our approach, we provide some visualization examples to illustrate the impact of our proposed method, as depicted in Fig.~\ref{fig:vis}. Instances marked with a red box signify inconsistency with the query image ID, while those marked with a green box indicate consistency.

These visualizations underscore the challenges of distinguishing certain images based solely on visual appearance. For instance, in the initial set of images (a), the individual in the third image is in a blue long-sleeved shirt and black pants, sharing a notable resemblance with the person in the query image.
In the second row of the visualization results, the woman in the first two images of (d) wears a black hoodie, denim pants, and black boots, and holds white rolls of paper, exhibiting a noticeable similarity in appearance with the person in the query image.
However, as can be seen from (f), the pedestrian in the correct images is facing away from the camera, and the coat does not exhibit any features of the white paper roll. 
In contrast, the person in the third image of (e) is attired in all black, and lacks a hat, but shares a dark hair color, bearing a strong resemblance to the pedestrians depicted in the two returned images. However, the stReID method fails to filter out this incorrect result based on the spatial-temporal statistic approach, and our method effectively filters out unreliable returned images.

\section{Conclusion}
\label{sec:conclusion}
The increasing volume of videos makes the traditional centralized ReID system impractical, and the current cloud-edge collaborative methods face challenges related to bandwidth constraints and search efficiency. To address these problems, we introduce a pioneering cloud-edge collaborative ReID framework. By leveraging a distribution-aware correlation modeling network, our approach enables efficient inference, ensuring the desired image returns to the cloud server as early as possible. 
Comparative experiments demonstrate our approach can reduce data transmission and improve the performance across various baselines, showing its superiority.
We also acknowledge that it requires time stamps and is inappropriate for mobile devices, which motivate our future research on utilizing unstable spatial-temporal data to achieve high-quality correlation learning.

\section*{Acknowledgments}
This work was supported in part by the Funds for the NSFC Project under Grant U24B20176, 62202063, 62406038, 62302058, and China Postdoctoral Science Foundation under Grant 2024M760280.
\bibliography{aaai25}

\end{document}